\pdfoutput=1

\documentclass[11pt]{article}

\usepackage[preprint]{acl}

\usepackage{times}
\usepackage{latexsym}
\usepackage{amsmath}
\usepackage{amssymb}

\usepackage[T1]{fontenc}

\usepackage[utf8]{inputenc}

\usepackage{microtype}

\usepackage{inconsolata}

\usepackage{booktabs}
\usepackage{multirow}
\usepackage{graphicx}
\usepackage{longtable}

\title{What Causes Knowledge Loss in Multilingual Language Models?}

\author{Maria Khelli$^1$, Samuel Cahyawijaya$^2$, Ayu Purwarianti$^1$, Genta Indra Winata$^3$ \\
  $^1$Institut Teknologi Bandung$\quad$$^2$Cohere$\quad^3$Capital One\\
  \texttt{khelli07.id@gmail.com, samuelcahyawijaya@cohere.com} \\
  \texttt{ayu@informatika.org, genta.winata@capitalone.com}}

\begin{document}
\maketitle
\begin{abstract}
Cross-lingual transfer in natural language processing (NLP) models enhances multilingual performance by leveraging shared linguistic knowledge. However, traditional methods that process all data simultaneously often fail to mimic real-world scenarios, leading to challenges like catastrophic forgetting, where fine-tuning on new tasks degrades performance on previously learned ones. Our study explores this issue in multilingual contexts, focusing on linguistic differences affecting representational learning rather than just model parameters. We experiment with 52 languages using LoRA adapters of varying ranks to evaluate non-shared, partially shared, and fully shared parameters. Our aim is to see if parameter sharing through adapters can mitigate forgetting while preserving prior knowledge. We find that languages using non-Latin scripts are more susceptible to catastrophic forgetting, whereas those written in Latin script facilitate more effective cross-lingual transfer.
\end{abstract}

\section{Introduction}
Cross-lingual transfer in natural language processing (NLP) models has demonstrated enhanced performance in multilingual contexts compared to monolingual settings, largely due to the advantages of leveraging cross-lingual knowledge \cite{hu2020xtreme, fitzgerald2023massive, winata2023nusax,winata2024worldcuisines}. Typically, training occurs only once simultaneously, where all available data is processed in a single training run. However, in real-world applications, data is often received sequentially over time, highlighting the importance of continuous model updates to maintain performance \cite{rolnick2019experience}. Unlike humans, who can retain prior knowledge while acquiring new skills, neural network models often struggle to preserve previously learned information when fine-tuned on new tasks, which known as catastrophic forgetting, a decline in performance on earlier tasks after the model is exposed to new data~\cite{winata2023overcoming}. To mitigate this issue, several studies have investigated continual learning strategies and the implementation of adapters \cite{badola2023parameter} as viable solutions. This limitation poses a significant challenge for multilingual NLP, as models must adapt to new languages while retaining previously acquired linguistic knowledge. Without an effective learning strategy, models risk performance degradation, rendering them less suitable for long-term deployment.

\begin{figure}[!t]
    \centering
    \includegraphics[width=0.49\textwidth]{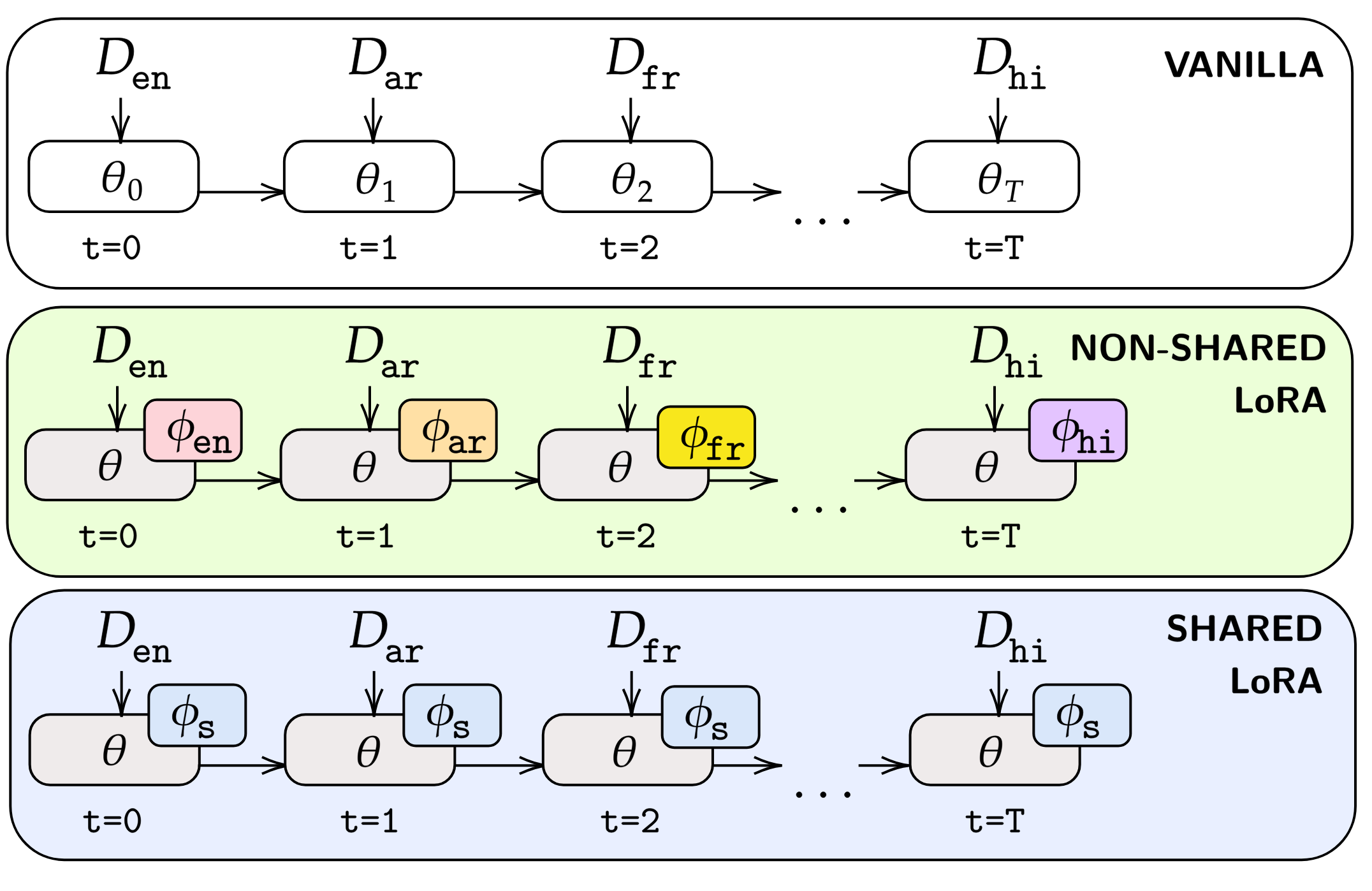}
    \caption{Pipeline for various approaches in lifelong learning. In our lifelong learning framework, we employ a LoRA-based approach where the parameters of the base model, denoted as $\theta$, remain fixed, and for VANILLA, the model parameters are updated at all times. We explore the phenomenon of multilingual knowledge loss by comparing the effects of training with both shared and non-shared parameters.}
    \label{fig:architecture}
\end{figure}

Lifelong learning is essential for integrating new annotated data across languages without requiring full retraining of systems. As language evolves and new data becomes available, models must adapt incrementally to minimize computational costs. This approach helps maintain efficiency and scalability, while addressing the challenge of catastrophic forgetting, which has been explored in various studies \cite{liu2021preserving, winata2023overcoming, badola2023parameter, mhamdi2023cross}. However, there is a lack of systematic analysis on this issue in multilingual contexts. This study aims to fill that gap by investigating factors contributing to catastrophic forgetting beyond model parameters, including how linguistic differences can affect representational learning and lead to knowledge erosion when learning multiple languages sequentially.


In this study, we investigate the effects of non-shared, partially shared, and fully shared parameters in a multilingual context, examining 52 languages through the use of LoRA adapters with varying ranks and different sharing model parameter settings as shown in Figure~\ref{fig:architecture}. Our primary focus is to assess the impact of parameter sharing on model performance, while also conducting a comprehensive analysis of the role that different languages play in catastrophic forgetting. Additionally, we explore sequential learning to identify when performance drops occur and whether these declines are influenced by the introduction of newly learned languages or the cumulative number of previously learned languages. Our contributions can be summarized as follows:
\begin{itemize}
    \item We examine the factors contributing to knowledge loss in multilingual language models, focusing on aspects such as language diversity, parameter sharing strategies, and base model selection within a lifelong learning framework for massively multilingual learning.
    \item We assess cross-lingual transferability and introduce multi-hop metrics to better understand the impact of language skills on model performance.
    \item We analyze model parameter adaptation to investigate trends in the model's ability to learn languages in a lifelong learning context.
\end{itemize}

\section{Methodology}

\subsection{Task Setup}

A sequence of \( T \) tasks is structured as an ordered set of datasets \( \mathcal{D} = \{D_1, D_2, \dots, D_t, \dots, D_T\} \), where each dataset \( D_t \) corresponds to a specific task \( t \), representing a distinct language. The model, parameterized by \( \theta_t \), undergoes iterative updates, with parameters at step \( t+1 \) being derived from those at step \( t \) through the function \( f(\theta_t, D_t) \). These updates are performed using gradient-based optimization to maximize the log-likelihood over dataset \( D_t \).

\subsection{Training Methods}

We use XLM-R$_\text{BASE}$~\cite{conneau2020unsupervised} as our base model and compare key methods with E5 instruct~\cite{wang2024multilingual} for evaluating the consistency of the findings. A classification layer is added on top of the encoder model, tailored sequence label of the slot filling. For adapter-based approaches, only the parameters within the adapter modules are updated during training.

\paragraph{\texttt{MULTI}.} A single model is trained on all languages simultaneously, optimizing over the entire dataset $\mathcal{D}$:
\begin{align}
    \theta_\text{MULTI} = \arg\max_{\theta} \sum_{t=1}^{T} \log p(D_t \mid \theta).
\end{align}
\paragraph{\texttt{MONO}.} Each language/task has its own independently trained model $\theta_t$:
\begin{align}
    \theta_t = \arg\max_{\theta} \log p(D_t \mid \theta), \\
    \forall t \in \{1, \dots, T\}.
\end{align}
\paragraph{\texttt{VANILLA}.} A single model is trained incrementally, updating parameters sequentially:
\begin{align}
    \theta_{t+1} \gets f(\theta_t, D_t), \forall t \in \{1, \dots, T-1\}.
\end{align}
\paragraph{\texttt{SHARED LoRA}.} A single LoRA adapter $\phi$ is trained while keeping the base model $\theta_0$ frozen:
\begin{align}
    \phi_{s} \gets f(\phi'_t, D_t), \theta = \theta_0, \forall t \in \{1, \dots, T-1\}.
\end{align}
\paragraph{\texttt{NON-SHARED LoRA}.} Each language has its own separate LoRA adapter $\phi_t$, while keeping the base model $\theta_0$ frozen:
\begin{align}
    \phi_t = \arg\max_{\phi} \log p(D_t \mid \theta_0, \phi), \forall t \in \{1, \dots, T\}.
\end{align}
The specific ordering of languages used in the \texttt{VANILLA} is specified in Appendix Table~\ref{tab:language-order}.

\subsection{Model Parameters Adaptation} 

We utilize low-rank adapters LoRA~\cite{hu2021lora} for training parameters to analyze the effectiveness to have sharing parameters. It is a parameter-efficient fine-tuning method for large pre-trained models leveraging the intrinsic low-dimensionality of parameter updates, reducing the need for full model adaptation. Instead of modifying dense layers directly, It freezes the pre-trained weights and introduces trainable low-rank matrices, significantly minimizing the number of learnable parameters and enhancing fine-tuning efficiency.

Formally, given a pre-trained weight matrix $W_0 \in \mathbb{R}^{d \times k}$, LoRA constrains the update $\Delta W$ to a low-rank decomposition:
\begin{equation}
\Delta W = BA,
\end{equation}
where $B \in \mathbb{R}^{d \times r}$ and $A \in \mathbb{R}^{r \times k}$, with rank $r \ll \min(d, k)$. This decomposition ensures that only $A$ and $B$ are updated while $W_0$ remains fixed. Consequently, the forward pass is expressed as:
\begin{equation}
h = W_0 x + \Delta W x = W_0 x + B A x,
\end{equation}
where $x$ is the input vector, and $h$ is the output. The low-rank update $\Delta W x$ is scaled by a constant factor $\frac{\alpha}{r}$, analogous to a learning rate, to regulate the magnitude of the update. LoRA offers key advantages: it enhances \textbf{memory} and \textbf{computational efficiency} by limiting trainable parameters, reducing resource requirements, and enabling modular fine-tuning. Its \textbf{linear structure} ensures \textbf{no additional inference latency}, while its \textbf{linear} property allows seamless integration. By leveraging low-rank adaptation, LoRA enables scalable and efficient model adaptation without compromising previously learned tasks.

\section{Experimental Setup}

\subsection{Datasets}

We utilize the MASSIVE, multilingual slot filling dataset \cite{fitzgerald2023massive}, which encompasses 52 languages and provides structured information, including scenarios, intents, utterances, and annotated utterances. Each language is uniformly represented, with 11.5K training samples, 2.03K validation samples, and 2.97K test samples.

\subsection{Hyper-parameters}

The training setup employed different configurations depending on whether LoRA was used. For models trained with LoRA, a learning rate of $5 \times 10^{-6}$ was applied, whereas models without LoRA used a higher learning rate of $5 \times 10^{-5}$. The number of training epochs is 100 for models with LoRA, and 50 for those without. Early stopping was implemented in both settings, with a patience of 15 epochs for LoRA and 5 epochs for non-LoRA models, based on the F1-score on validation data. The LoRA configuration included a dropout rate of 0.1, and the scaling factor $\alpha$ was set equal to the rank (32, 64, 256 respectively).

\subsection{Evaluation Metrics}

We evaluate the performance of the model using average F1 score for the learned tasks and visualized its progression over number of learned languages, as illustrated in Figure~\ref{fig:xlmr-result}. Besides that, there are additional metrics, particularly for sequential methods such as \texttt{VANILLA} and \texttt{SHARED LoRA}. 

\subsubsection{Performance Shift}
This metrics is used to measure the average performance shift, which quantifies the change in a previously learned language performance after training in a new language. Formally, we define the average performance change as follows:  
\begin{align}  
\mathcal{P}_{\text{avg}} = \frac{1}{N} \sum_{n=1}^{N} (\mathcal{P}_{t} - \mathcal{P}_{t+1}),  
\end{align}  
where \( \mathcal{P}_{t} \) and \( \mathcal{P}_{t+1} \) represent the average F1 score over all previously encountered tasks at time steps \( t \) and \( t+1 \), respectively. To account for variability in task sequences, the performance changes are averaged over five times ($N=5$).

\subsection{Cross-lingual Transfer}

We assess cross-lingual transfer effectiveness using Cross-lingual Forward Transfer (CFT) and Cross-lingual Backward Transfer (CBT) metrics from \citet{winata2023overcoming} and we introduce a new metric, Multi-Hop Forward Transfer (MFT), and Multi-Hop Backward Transfer (MBT) to measure the multi-hop transfer for each language. Let $R \in \mathbb{R}^{T \times T}$ be a matrix where $R_{i,j}$ represents the test score performance on task $t_j$ after training on the last sample from task $t_i$. The two types of metrics are defined as follows.

\paragraph{Cross-lingual Forward Transfer (CFT).} The metric evaluates the model’s ability to generalize to unseen languages by assessing test performance on tasks not encountered during training. It is defined as:
\begin{equation}
    CFT = \frac{1}{T - 1} \sum_{i=1}^{T-1} \bar{X}_i,
\end{equation}
where
\begin{equation}
    \bar{X}_i = \frac{1}{T - i} \sum_{j=i+1}^{T} R_{i,j}.
\end{equation}
Here, $\bar{X}_i$ represents the average performance across future tasks ($t_j > t_i$).

\paragraph{Cross-lingual Backward Transfer (CBT).} The metric measures the impact of learning a new task $t_i$ on the performance of previously learned tasks. It is formally defined as:
\begin{equation}
    CBT = \frac{1}{T - 1} \sum_{i=1}^{T-1} \left( R_{T-1,i} - R_{i,i} \right).
\end{equation}
This metric quantifies the extent of catastrophic forgetting, where adding a new task may negatively impact the performance of past tasks.

\paragraph{Multi-Hop Forward Transfer (MFT).} The metric measures the knowledge transfer effect between tasks separated by multiple learning steps. For a hop distance $h$, MFT is defined as:
\begin{equation}
MFT_h = \frac{1}{|L|} \sum_{l \in L} (\mathcal{P}_{i+h} - \mathcal{P}_{i-1}),
\end{equation}
where $\mathcal{P}i$ represents the average performance on tasks seen up to step $i$. This metric quantifies how learning a language affects performance on another language that will be encountered $h$ steps later in the training sequence.

\paragraph{Multi-Hop Backward Transfer (MBT).} The metric similarly evaluates the effect of learning a new task on the performance of tasks encountered several steps earlier. For a hop distance $h$, MBT is defined as:
\begin{equation}
MBT_h = \frac{1}{|L|} \sum_{l \in L} (\mathcal{P}_{i} - \mathcal{P}_{i-h-1}).
\end{equation}
This metric measures how training on a language affects the performance on languages that were learned $h$ steps before in the training sequence. The term \textit{multi-hop} refers to our evaluation across multiple hops, as illustrated in Figure~\ref{fig:hop-backward}. A hop distance of zero corresponds to the performance change metric.

\begin{table}[!t]
    \centering
    \resizebox{.49\textwidth}{!}{  
    \begin{tabular}{lrr|rrr}  
        \toprule
        \textbf{Method} & \textbf{Params (M)} & \textbf{F1 (\%)} & \multicolumn{3}{c}{\textbf{Language Vitality}} \\
        & & & \textbf{Low} & \textbf{Mid} & \textbf{High} 
        \\
        \midrule
        MULTI & 278.04  & \textbf{75.03} & 75.42 & \textbf{75.84} & 72.63 \\
        $\quad$$r = $ 32 & 5.36  & 74.19 & 74.27 & \textbf{75.17} & 71.83 \\
        $\quad$$r = $ 64 & 10.72  & 73.79 & 74.00 & \textbf{74.56} & 71.73 \\
        $\quad$$r = $ 256 & 42.86  & 74.11 & 74.16 & \textbf{74.83} & 72.41\\
        MONO & 14,458.27  & 72.98 & 73.66 & \textbf{74.11} & 69.43 \\
        VANILLA & 278.04  & 66.16 & 65.70 & \textbf{67.65}& 63.46 \\
        \midrule
        \multicolumn{2}{l}{SHARED LoRA} & & \\
        $\quad$$r = $ 32 & 5.36 & 60.24 & 59.35 & \textbf{62.34} & 56.75 \\
        $\quad$$r = $ 64 & 10.72  &\textbf{61.26}& 60.55 & \textbf{63.37} & 57.48 \\
        $\quad$$r = $ 256 & 42.86  & 60.16 & 59.06 & \textbf{62.15} & 57.22 \\
        \midrule
        \multicolumn{2}{l}{NON-SHARED LoRA} & & \\
        $\quad$$r = $ 32 & 278.04 & 72.14 & 72.42 & \textbf{73.39} & 68.89\\
        $\quad$$r = $ 64 & 557.19  & 72.38 & 72.55 & \textbf{73.48} & 69.65\\
        $\quad$$r = $ 256 & 2,228.75  & \textbf{73.16} & 73.82 & \textbf{74.26} & 69.73\\
        \bottomrule
    \end{tabular}
    }
    \caption{Comparison of methods based on trainable parameters (in million parameters) and averaged F1 (\%). Lower trainable parameters is better, higher average performance is better.}
    \label{tab:comparison}
\end{table}

\begin{figure}[!t]
    \includegraphics[width=0.48\textwidth]{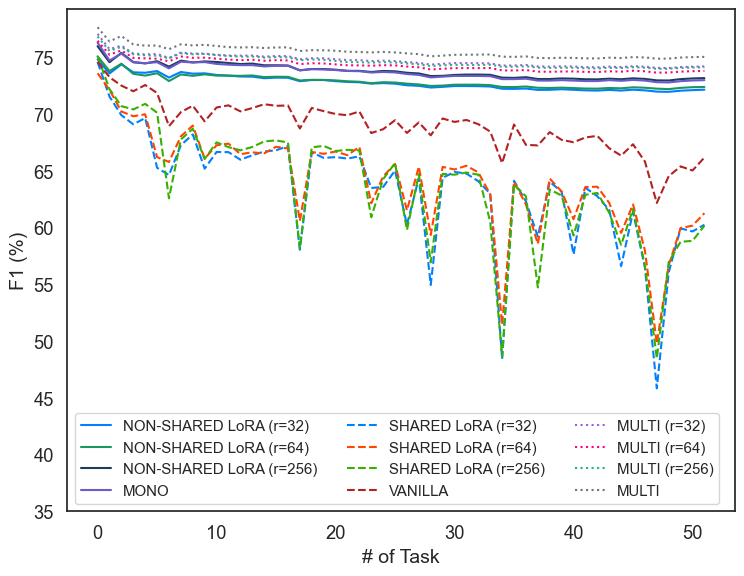} 
    \caption{Performance results after training each language over the time.}
    \label{fig:xlmr-result}
\end{figure}

\section{Results}

\begin{figure*}[!t]
    \includegraphics[width=\textwidth]{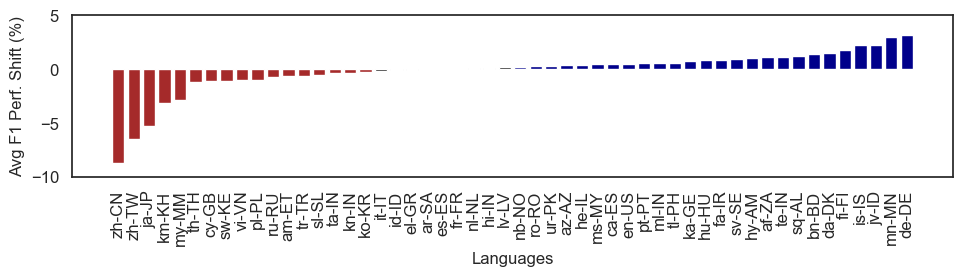} 
    \caption{Performance change after training a certain language on x-axis in sequential training (\texttt{VANILLA}). Chinese \texttt{(zh-CN)} exhibits the most significant performance decline, while German \texttt{(de-DE)} serves as the most effective donor language, enhancing overall performance.}
    \label{fig:performance-impact}
\end{figure*}

Figure~\ref{fig:performance-impact} illustrates the impact of training different languages sequentially on model performance towards learned language, measured by the average F1 change across 5 different orders. 

\begin{figure*}[!th]
    \centering
    \includegraphics[width=.8\textwidth]{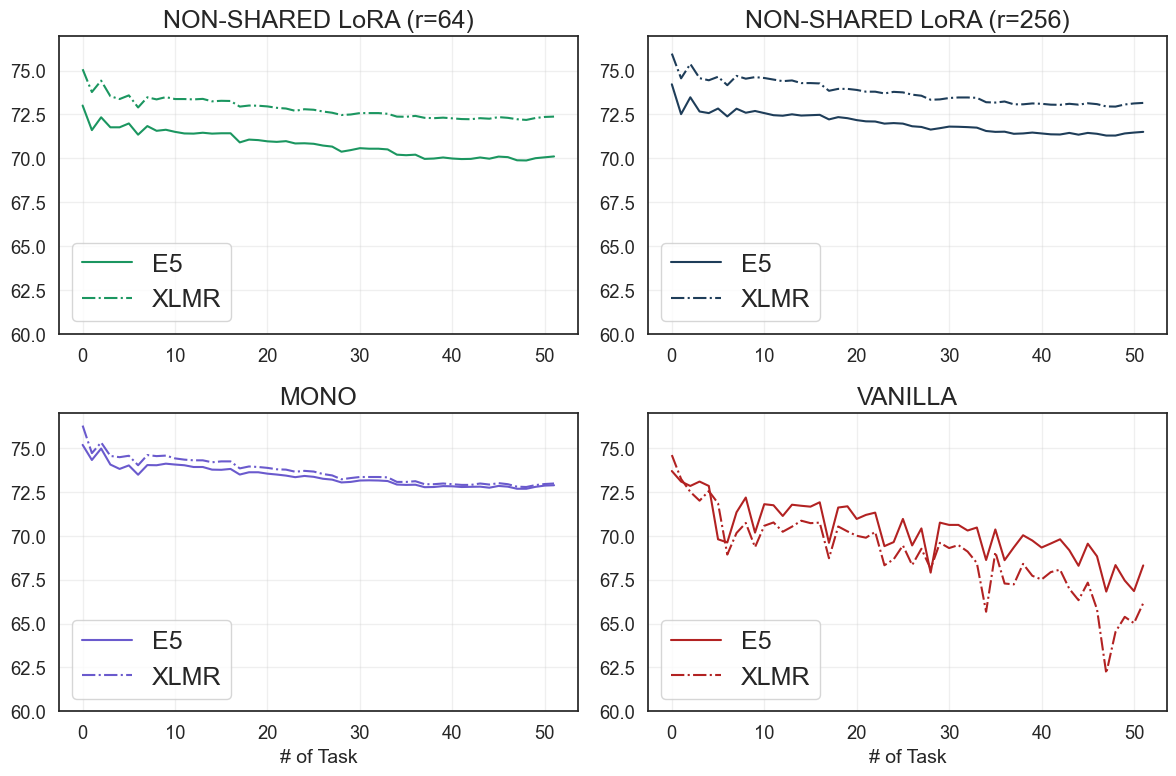} 
    \caption{Comparison results between XLM-R and E5 models.}
    \label{fig:e5-vs-xlmr}
\end{figure*}

\paragraph{Performance vs. Model Parameters.}
Table~\ref{tab:comparison} presents a comparison of training methods in terms of average F1 score and trainable parameters. The \texttt{MULTI} method achieves the best overall performance (75.03\%) with a moderate parameter footprint (278.04M), offering an excellent balance between effectiveness and efficiency. On the opposite end, \texttt{MONO}, which trains an entirely separate model per language, consumes an enormous parameter budget (14{,}458.27M) while yielding only moderate performance (72.98\%), highlighting the inefficiency of isolated training.

Among parameter-efficient alternatives, LoRA-based approaches exhibit varying trade-offs. \texttt{NON-SHARED LoRA} performs competitively (up to 73.16\% at rank 256), benefiting from task-specific specialization, albeit with moderate parameter cost (2{,}228.75M). In contrast, \texttt{SHARED LoRA}'s best result dramatically reduces the number of trainable parameters (e.g., 10.72M at rank 256) but suffers heavily in performance, dropping to as low as 61.26\%.

Crucially, increasing the LoRA rank—while expanding the model's capacity—does not substantially improve performance. For instance, \texttt{MULTI} with rank 32 (74.19\%) performs nearly as well as at rank 256 (74.11\%), and similar diminishing returns are observed across both \texttt{SHARED} and \texttt{NON-SHARED LoRA}. This trend extends to transfer metrics: Table~\ref{tab:cbt-cft} shows that higher rank under \texttt{SHARED LoRA} does not significantly improve forward transfer—CFT remains within the narrow band of $0.51$--$0.53$. These results highlight a key trade-off: higher trainable parameters generally improve performance, but the efficiency of parameter usage varies across methods. The \texttt{MULTI} method provides the best balance between parameter efficiency and performance, while LoRA-based approaches demonstrate potential for parameter-efficient training at the cost of reduced performance. However, it should be noted that the \texttt{MULTI} method might not be trainable in parallel like the \texttt{NON-SHARED LoRA} method. Hence, in some scenarios, the \texttt{NON-SHARED LoRA} method should be considered.

\paragraph{Trends Between Models.} Figure~\ref{fig:xlmr-result} illustrates how different training strategies affect performance over time. A key trend is that \texttt{MULTI} methodA (dotted lines), trained jointly on all languages, exhibit consistent performance, maintaining F1-scores above 73\% throughout training. In contrast, sequential learning models show clear signs of degradation as training progresses. The \texttt{VANILLA} model suffers from moderate catastrophic forgetting, with F1-score reductions of 10--15 points. \texttt{SHARED LoRA} fares worse, degrading by as much as15--30 points across tasks. Meanwhile, \texttt{NON-SHARED LoRA} offers more stable performance across steps, ranging between 70--73\% and demonstrating greater resilience to forgetting.

These observations are further supported by Table~\ref{tab:cbt-cft}, which reports backward and forward transfer scores. The \texttt{VANILLA} model achieves a CBT of $\mathbf{-0.08}$ and CFT of $\mathbf{0.55}$, suggesting that while it suffers from forgetting, it still generalizes reasonably well to future tasks. \texttt{SHARED LoRA}, however, shows consistently more negative CBT scores ($-0.13$ to $-0.14$), confirming its vulnerability to catastrophic forgetting. This performance is also reflected in CFT, where the scores are also lower than \texttt{VANILLA} method. Together, these findings underscore the importance of balancing task generalization and knowledge retention, particularly in continual cross-lingual setups.

\begin{table}[!th]
    \centering
    \resizebox{.36\textwidth}{!}{
    \begin{tabular}{lrr}  
        \toprule
        \textbf{Method} & \textbf{CBT} & \textbf{CFT} \\
        \midrule
        VANILLA & \textbf{-0.08} & \textbf{0.55} \\
        SHARED LoRA \\
        $\quad$$r = $ 32 & -0.13 & 0.52 \\
        $\quad$$r = $ 64 & -0.12 & 0.53 \\
        $\quad$$r = $ 256 & -0.14 & 0.51 \\
        \bottomrule
    \end{tabular}
    }
    \caption{CBT and CFT metrics for VANILLA and SHARED LoRA models — higher values indicate better performance.}
    \label{tab:cbt-cft}
\end{table}

\paragraph{Comparison XLM-R and E5 Models.} Figure~\ref{fig:e5-vs-xlmr} presents a comparison of XLM-R and E5 models across different training methods. Despite variations in methodology, the general pattern of results remains consistent across models. Overall, XLM-R performs better than E5, except in \texttt{VANILLA} method where E5 tends to outperform  XLM-R$_\text{BASE}$, though performance degradation due to forgetting is still evident. The results suggest that while different methods and model architectures influence the degree of forgetting, the overall trend of performance degradation remains a common characteristic across all settings.

\section{Analysis on Languages}

To frame our analysis, we interpret MFT as measuring a language's ability to \textbf{donate} knowledge to subsequent languages, while MBT reflects how well a language \textbf{receives} and retains knowledge after subsequent training steps. This donor-receiver perspective allows us to reason about asymmetries in cross-lingual transfer.

\subsection{Languages Affect Forgetting}

\begin{figure*}[!th]
    \centering
    \includegraphics[width=1\textwidth]{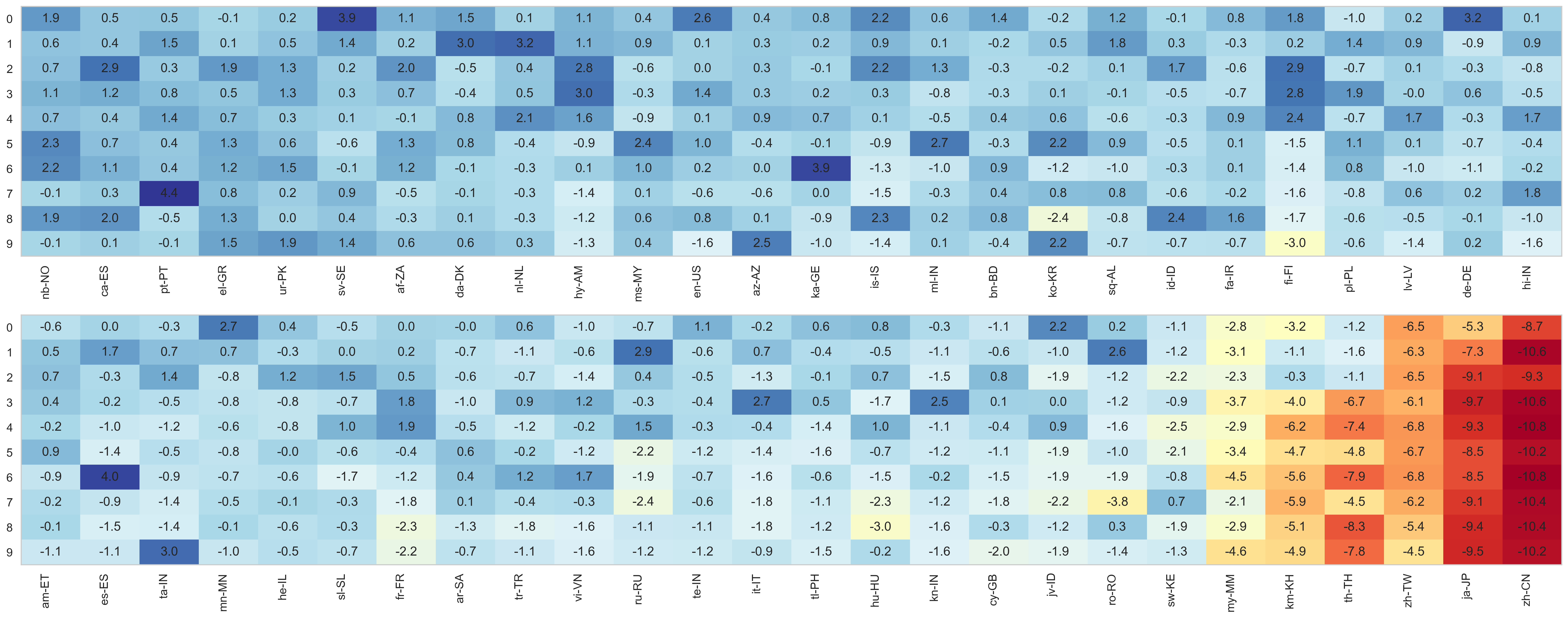} 
    \caption{Heatmap of Multi-hop Backward Transfer (MBT), illustrates how training on later languages affects earlier ones over increasing hop distances (rows 0–9). Cooler colors indicate positive backward transfer, while warmer colors reflect degradation in performance. Notable negative effects are observed in languages like ja-JP, zh-CN, and zh-TW.}
    \label{fig:hop-backward}
\end{figure*}

The results reveal that certain languages significantly impact the model’s capacity to retain prior knowledge. Training on \texttt{Chinese (zh-CN)}, \texttt{Japanese (ja-JP)}, and \texttt{Traditional Chinese (zh-TW)} consistently leads to the most pronounced cases of catastrophic forgetting. This is evidenced by their strongly negative MBT values in Figure~\ref{fig:hop-backward} and severe performance degradation in Figure~\ref{fig:performance-impact}, particularly when these languages are introduced later in the training sequence. As receivers, these languages appear highly vulnerable to interference from prior tasks. More detailed explanation can be seen in Appendix. In contrast, languages such as \texttt{Norwegian (nb-NO)}, \texttt{Catalan (ca-ES)}, \texttt{Portuguese (pt-PT)}, and \texttt{Greek (el-GR)} show some of the highest MBT scores across hop distances. These languages maintain stability when trained after others and also preserve prior task performance, indicating they are robust receivers. Interestingly, they may also act as indirect donors by not interfering with earlier knowledge. 

However, not all performance trends align perfectly with MBT. For example, \texttt{German (de-DE)} appears beneficial in performance drop metrics (Figure~\ref{fig:performance-impact}), yet does not rank highly in MBT. This suggests that its apparent advantage may be due to its position in the training sequence—e.g., being trained before high-forgetting languages—rather than any inherent ability to preserve earlier knowledge. This underscores an important point: interpreting language influence solely through performance drop can be misleading. MBT offers a more principled, sequence-agnostic perspective on which languages genuinely aid in preserving prior knowledge and resisting catastrophic forgetting.

\begin{figure*}[!th]
    \centering
    \includegraphics[width=1\textwidth]{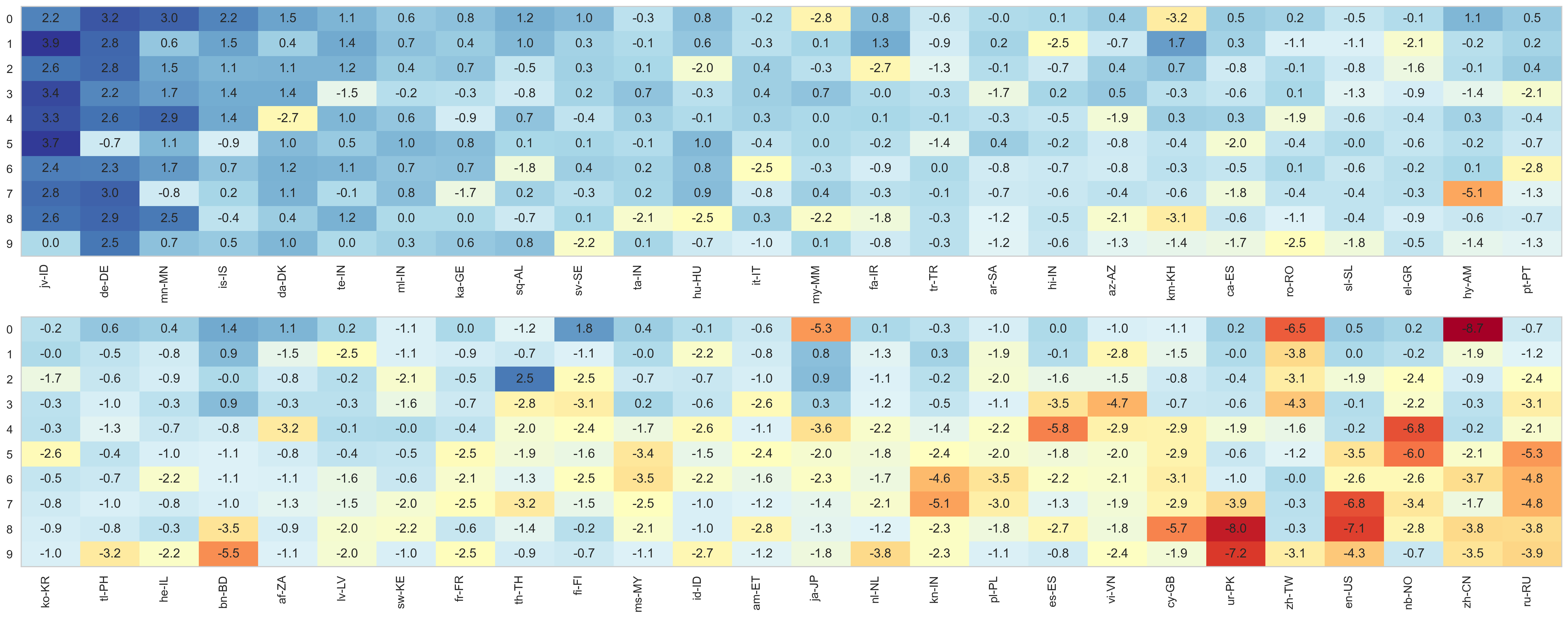} 
    \caption{Heatmap of Multi-hop Forward Transfer (MFT), represents each language's ability to donate knowledge to subsequent tasks. Higher values (bluer) indicate stronger positive transfer, while lower values (yellower to redder) reflect limited or negative influence on future learning.}
    \label{fig:hop-forward}
\end{figure*}

\subsection{Latin vs. Non-Latin Scripts}

Script similarity plays a significant role in cross-lingual knowledge transfer. In both MFT and MBT heatmaps, we observe that languages using Latin scripts—such as \texttt{es-ES}, \texttt{fr-FR}, and \texttt{de-DE}—tend to be strong donors and stable receivers. They benefit more from training on other languages and also suffer less from catastrophic forgetting. This likely reflects greater subword token overlap and lexical similarity, which help preserve learned representations under shared tokenization.

In contrast, non-Latin script languages, especially those using logographic (e.g., \texttt{zh-CN}) or abugida scripts (e.g., \texttt{th-TH}, \texttt{hi-IN}), tend to be weak donors and vulnerable receivers. These languages show low MFT—suggesting limited forward transfer to other tasks—and highly negative MBT, indicating susceptibility to forgetting. The subword tokenizer, likely optimized for Latin-based alphabets, aggravates this imbalance. This highlights a fundamental challenge for multilingual continual learning: shared vocabulary spaces can lead to representational dominance of Latin-script languages, marginalizing others.

\subsection{Language Family}

While language family information is not explicitly modeled, typologically or lexically similar languages often demonstrate mutual reinforcement in transfer. Under the donor-receiver lens, we observe that Romance languages such as \texttt{es-ES}, \texttt{pt-PT}, and \texttt{fr-FR} frequently act as strong donors (high MFT) and reliable receivers (stable MBT), especially when positioned near each other in the training sequence. Similarly, Germanic languages like \texttt{nl-NL}, \texttt{sv-SE}, and \texttt{de-DE} show stable transfer interactions.

However, these patterns are not universal. The apparent family-related benefits may arise from shared scripts and vocabulary rather than deep structural similarity. For instance, several Indo-European languages from different branches perform well together, likely due to orthographic overlap. Conversely, languages from distant families—such as Sino-Tibetan (\texttt{zh-CN}), Austroasiatic (\texttt{km-KH}), or Afro-Asiatic (\texttt{ar-SA})—often act as poor receivers (low MBT) and limited donors (low MFT), especially when sequenced after typologically dissimilar languages. Future work could explicitly incorporate phylogenetic distances to better disentangle the impact of language family on multilingual continual learning.

\subsection{Language Vitality}

Language vitality—encompassing speaker population, data availability, and digital presence—also plays a nuanced role in continual learning dynamics. As receivers, high-vitality languages such as \texttt{zh-CN}, \texttt{ja-JP}, and \texttt{hi-IN} show some of the most negative MBT scores, indicating that they are especially vulnerable to forgetting. Surprisingly, they also make relatively poor donors, as reflected in lower MFT scores compared to more typologically compatible mid-vitality languages.

This counterintuitive trend is clarified in Table~\ref{tab:comparison}, where mid-vitality languages consistently achieve the highest F1 scores across model variants. These languages appear to strike a balance: they share enough structure with other languages to act as effective donors, while remaining resilient as receivers under sequential training. In contrast, high-vitality languages—despite abundant resources—struggle under parameter-efficient continual learning setups. Their unique token distributions and structural divergence make them harder to adapt to and easier to overwrite. These findings suggest that vitality-aware scheduling or modularization may be critical for improving cross-lingual robustness in continual learning scenarios.

\section{Related Work}

Catastrophic forgetting is a significant challenge in neural networks, where models lose previously acquired knowledge when fine-tuned on new tasks \cite{McCloskey1989}. This issue is particularly pronounced in multilingual contexts, as models must adapt to new languages without degrading performance on previously learned ones \cite{winata2023overcoming}. To mitigate this, various strategies have been proposed, including memory replay \cite{rolnick2019experience}, regularization techniques \cite{kirkpatrick2017overcoming}, and architectural innovations like progressive networks \cite{rusu2016progressive}. 

Lifelong learning also known as continual learning, is an emerging approach that enables models—particularly LLMs and their agents—to continuously acquire new knowledge while retaining prior capabilities. This knowledge can be integrated into LLMs either by updating model parameters through training or adapters, or by leveraging external sources like Wikipedia or tools without modifying the model itself or knowledge base~\cite{zheng2024towards}. Recent work extends lifelong learning to agent-based settings, decomposing it into perception, memory, and action modules that together support continuous adaptation \cite{zheng2025lifelong}.

For internal knowledge updates, adapters have proven to be a lightweight and effective solution, introducing small, task-specific modules that can be fine-tuned independently, reducing interference across tasks \cite{houlsby2019parameter,winata2021adapt,hu2021lora}. The MAD-X framework \cite{pfeiffer2020madx} enhances cross-lingual transfer by separating language and task adaptation, while language-specific adapters balance specialization and sharing \cite{badola2023parameter}. Additionally, methods like AdapterFusion \cite{pfeiffer2020adapterfusion} combines task-specific adapters through a learned composition layer, promoting parameter sharing and effective transfer learning while minimizing forgetting.

\section{Conclusion}
Our paper highlights the critical challenges of catastrophic forgetting in cross-lingual transfer for multilingual NLP models with 52 languages. We provide insights into how various parameter-sharing strategies can influence knowledge retention and overall model performance. Our findings indicate that partial parameter sharing can effectively mitigate forgetting while maintaining performance, presenting a promising approach for developing more robust multilingual NLP systems. Additionally, we identify that certain languages during training can negatively impact performance, contributing to catastrophic forgetting. Overall, this research enhances the ongoing efforts to improve the adaptability and efficiency of NLP models in real-world NLP applications.

\section*{Limitations}
In this paper, we concentrate our investigation on XLM-R model and use E5, rather than exhaustively evaluating every possible model due to resource constraints. This focused approach allows us to provide a more in-depth analysis of these models and their performance in cross-lingual contexts.

\section*{Ethical Considerations}
In our evaluation of language models for multilingual tasks, we place strong emphasis on transparency and fairness. We carefully design and document our data collection and evaluation methodologies to ensure they are consistent, unbiased, and reproducible. By applying uniform assessment criteria across models, we aim to enable meaningful and equitable comparisons.

\bibliography{custom}

\appendix


\begin{table*}[!th]
    \centering
    \resizebox{\textwidth}{!}{
    \begin{tabular}{cl}
    \toprule
    \textbf{Order} & \textbf{Languages in ISO 639-1} \\
    \midrule
    1 & en-US, ru-RU, id-ID, vi-VN, fa-IR, th-TH, ja-JP, de-DE, ro-RO, hu-HU, fr-FR, fi-FI, ko-KR, es-ES, pt-PT, nb-NO, el-GR, \\
      & zh-CN, da-DK, pl-PL, he-IL, it-IT, nl-NL, ar-SA, tr-TR, hi-IN, zh-TW, ta-IN, sv-SE, sl-SL, ca-ES, ka-GE, lv-LV, ms-MY, bn-BD, \\
      & ml-IN, az-AZ, ur-PK, hy-AM, sq-AL, te-IN, kn-IN, is-IS, tl-PH, mn-MN, my-MM, sw-KE, km-KH, af-ZA, am-ET, cy-GB, jv-ID \\
    \midrule
    2 & tr-TR, ro-RO, ur-PK, es-ES, hi-IN, pl-PL, hy-AM, sv-SE, sl-SL, ta-IN, te-IN, ml-IN, id-ID, ka-GE, el-GR, ko-KR, de-DE, \\
      & fa-IR, ms-MY, ca-ES, az-AZ, nl-NL, pt-PT, fr-FR, hu-HU, sw-KE, mn-MN, he-IL, zh-CN, fi-FI, ru-RU, is-IS, cy-GB, ja-JP, sq-AL, \\
      & vi-VN, th-TH, jv-ID, it-IT, my-MM, kn-IN, lv-LV, am-ET, nb-NO, ar-SA, en-US, af-ZA, zh-TW, bn-BD, da-DK, km-KH, tl-PH \\
    \midrule
    3 & sv-SE, nl-NL, fi-FI, kn-IN, hu-HU, ms-MY, es-ES, my-MM, is-IS, ko-KR, af-ZA, vi-VN, bn-BD, tr-TR, tl-PH, lv-LV, ru-RU, fr-FR, \\
      & en-US, ro-RO, am-ET, he-IL, hi-IN, ja-JP, te-IN, id-ID, ta-IN, it-IT, jv-ID, nb-NO, ka-GE, sq-AL, ca-ES, az-AZ, zh-TW, fa-IR, \\
      & mn-MN, zh-CN, de-DE, da-DK, ml-IN, sw-KE, sl-SL, km-KH, ar-SA, pt-PT, cy-GB, ur-PK, hy-AM, el-GR, pl-PL, th-TH \\
    \midrule
    4 & nb-NO, ta-IN, th-TH, fi-FI, ru-RU, af-ZA, vi-VN, ko-KR, ro-RO, km-KH, is-IS, ms-MY, sl-SL, en-US, hi-IN, he-IL, bn-BD, \\
      & pt-PT, fa-IR, sv-SE, am-ET, kn-IN, az-AZ, tl-PH, ar-SA, nl-NL, cy-GB, hy-AM, it-IT, de-DE, da-DK, te-IN, hu-HU, lv-LV, \\
      & zh-CN, mn-MN, es-ES, ca-ES, pl-PL, fr-FR, ja-JP, ka-GE, sw-KE, id-ID, zh-TW, jv-ID, sq-AL, el-GR, tr-TR, my-MM, ml-IN, ur-PK \\
    \midrule
    5 & mn-MN, ml-IN, is-IS, fa-IR, az-AZ, pl-PL, de-DE, ko-KR, ar-SA, sw-KE, jv-ID, sq-AL, tl-PH, ru-RU, lv-LV, fr-FR, ro-RO, \\
      & ka-GE, cy-GB, tr-TR, he-IL, sl-SL, af-ZA, nl-NL, my-MM, hu-HU, hi-IN, vi-VN, it-IT, pt-PT, da-DK, ca-ES, am-ET, el-GR, ta-IN, \\
      & id-ID, te-IN, sv-SE, bn-BD, ur-PK, en-US, kn-IN, ms-MY, nb-NO, es-ES, fi-FI, zh-TW, zh-CN, ja-JP, th-TH, km-KH, hy-AM \\
    \bottomrule
    \end{tabular}
    }
    \caption{Language orders in the sequential training experiments.} \label{tab:language-order}
\end{table*}

\begin{figure*}[!t]
    \centering
    \includegraphics[width=\linewidth]{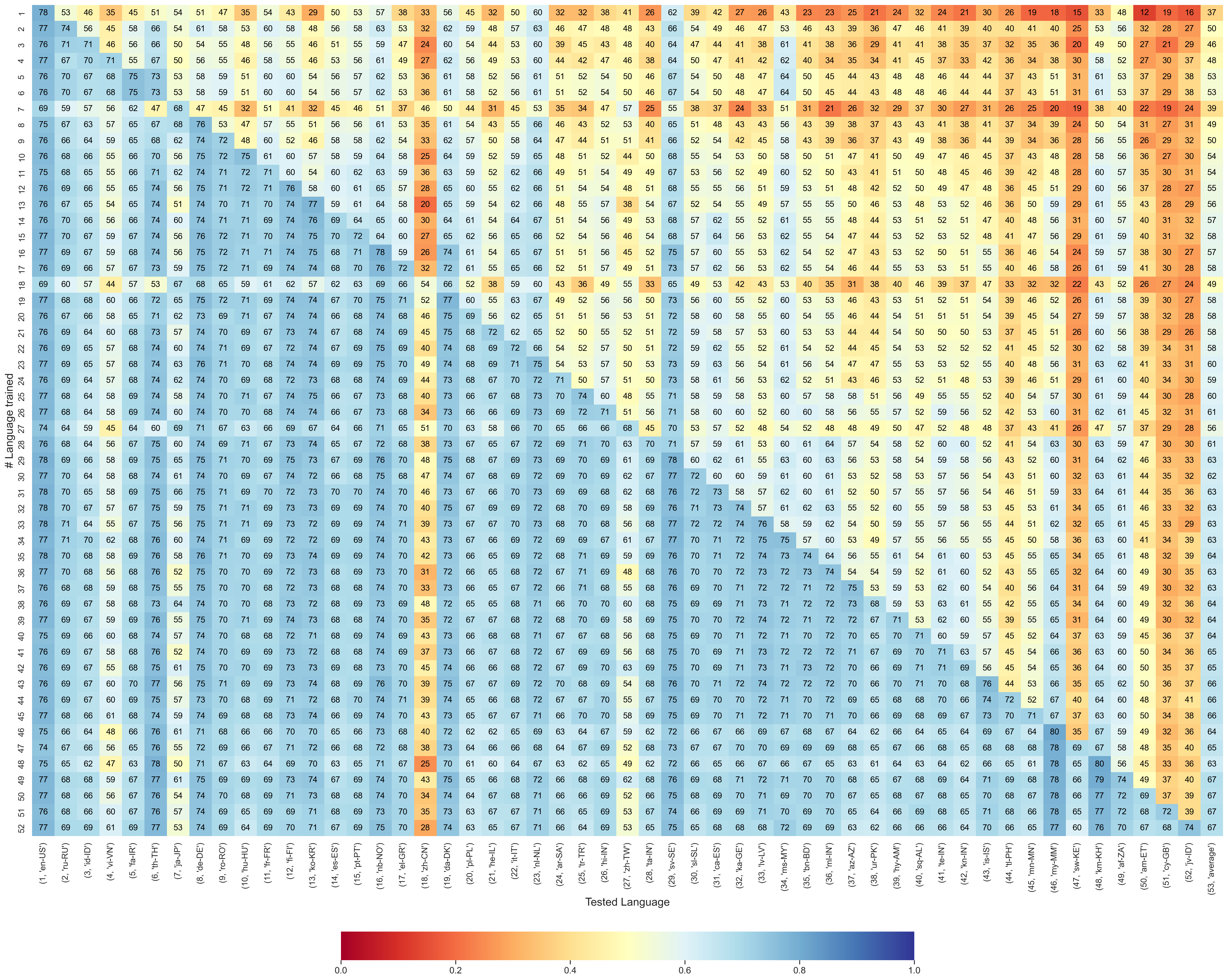} 
    \caption{Heatmap on \texttt{VANILLA} method for first language order.}
    \label{fig:heatmap-nolora}
\end{figure*}

\section{Detailed Results}
\label{sec:appendix}

\subsection{Language Order}
Table~\ref{tab:language-order} presents the language orders used in the sequential training experiments. These orders are used to train models in a step-by-step fashion, where each iteration introduces a new language. The results from these training sequences are subsequently used to compute aggregate metrics, as shown in Figure~\ref{fig:xlmr-result} and Figure~\ref{fig:e5-vs-xlmr}.

The first order is derived based on the amount of language resources available in the XLM-R model \cite{conneau2020unsupervised}. This order reflects the relative training data size used during XLM-R’s pretraining, with high-resource languages appearing earlier in the sequence. The remaining orders (2 through 5) are randomly shuffled variants to introduce diversity and reduce potential order bias. However, in the fifth order, languages that are found to be particularly destructive—i.e., those that tend to cause performance degradation on previously learned languages—are deliberately placed toward the end of the sequence. This design allows us to analyze how the position of destructive languages affects knowledge retention and transfer in sequential multilingual training.

\subsection{Heatmap on \texttt{VANILLA} method for first language order}
The heatmap on Figure~\ref{fig:heatmap-nolora} provides a detailed visualization of the model’s performance across training iterations (represented by rows) and evaluated languages (represented by columns). In each iteration, the model is trained on a new language. For instance, as shown in the figure, the first iteration trains on \texttt{en-US}, the second on \texttt{ru-RU}, the third on \texttt{id-ID}, and so forth. After training on a language, the model's performance on that language typically improves. This trend is reflected in the heatmap: the lower-left triangle (below the diagonal), corresponding to previously learned languages, tends to display cooler colors, indicating better performance; in contrast, the upper-right triangle (unlearned languages) often exhibits warmer colors, reflecting performance degradation. 

This visualization clearly highlights cross-lingual interactions—specifically, how training on a new language can either benefit or harm performance on other languages. For example, in row 18, where the model is trained on \texttt{zh-CN}, the corresponding row becomes noticeably warmer compared to previous iterations, suggesting a general decline in performance across many languages. However, for linguistically related languages such as \texttt{ja-JP}, performance actually improves. This indicates that while \texttt{zh-CN} introduces interference for many languages, it serves as a helpful donor for \texttt{ja-JP}, likely due to shared linguistic representations, such as the incorporation of Chinese characters in the Japanese writing system.

\end{document}